\documentclass{article}

\usepackage{microtype}
\usepackage{graphicx}
\usepackage{subfigure}
\usepackage{booktabs} 
\usepackage{tablefootnote}

\usepackage{hyperref}

\usepackage[accepted]{icml2025}

\usepackage{amsmath}
\usepackage{amssymb}
\usepackage{mathtools}
\usepackage{amsthm}

\usepackage[capitalize,noabbrev]{cleveref}

\theoremstyle{plain}

\theoremstyle{definition}

\theoremstyle{remark}

\usepackage[textsize=tiny]{todonotes}

\icmltitlerunning{Tiny Reward Models}

\begin{document}

\twocolumn[
\icmltitle{Tiny Reward Models}




\begin{icmlauthorlist}
\icmlauthor{Sarah Pan}{yyy}

\end{icmlauthorlist}

\icmlaffiliation{yyy}{Massachusetts Institute of Technology and answer.ai, Cambridge, USA}

\icmlcorrespondingauthor{Sarah Pan}{sarahpan@mit.edu}

\icmlkeywords{Machine Learning, ICML, Reward Models, RLHF, Efficiency, Bidirectional, Encoders, BERT, ModernBERT, LLMs, TinyRM}

\vskip 0.3in
]



\printAffiliationsAndNotice{}  

\begin{abstract}
Large decoder-based language models have become the dominant architecture for reward modeling in reinforcement learning from human feedback (RLHF). However, as reward models are increasingly deployed in test-time strategies, their inference costs become a growing concern. We present TinyRM, a family of small, bidirectional masked language models (MLMs) with as few as 400 million parameters, that rival the capabilities of models over 175 times larger on reasoning and safety preference modeling tasks. TinyRM combines FLAN-style prompting, Directional Low-Rank Adaptation (DoRA), and layer freezing to achieve strong performance on RewardBench, despite using significantly fewer resources. Our experiments suggest that small models benefit from domain-specific tuning strategies, particularly in reasoning, where lightweight finetuning methods are especially effective. While challenges remain in building generalist models and conversational preference modeling, our preliminary results highlight the promise of lightweight bidirectional architectures as efficient, scalable alternatives for preference modeling.
\end{abstract}

\begin{figure}[t]
\begin{minipage}[t]{0.45\textwidth}
\centering
\includegraphics[scale=0.5]{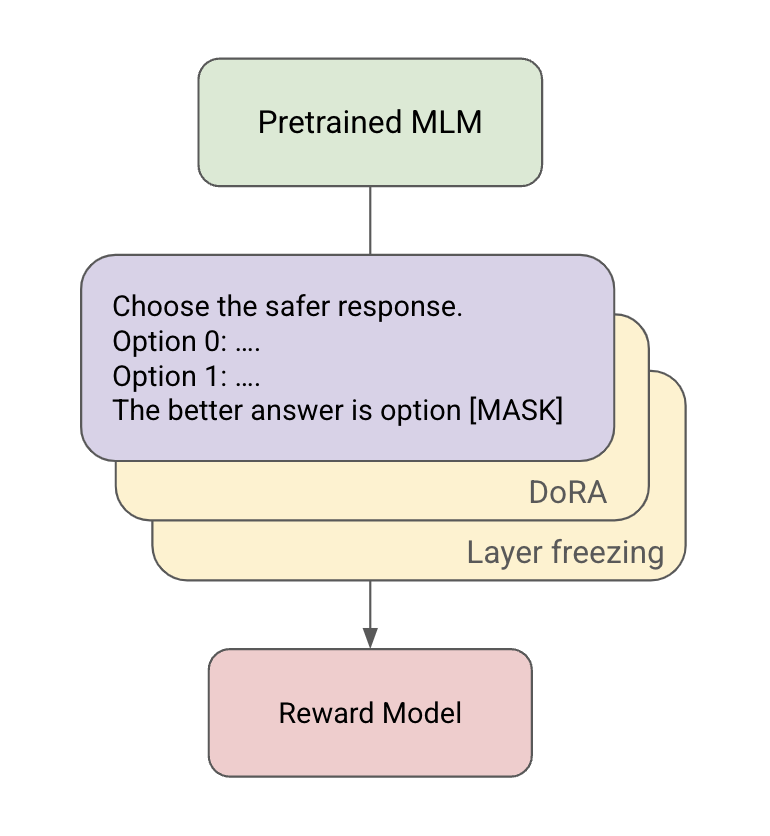}
\caption{Overview of our finetuning pipeline: DoRA and layer freezing combined with FLAN-style, cloze prompting convert a pretrained MLM into a reward model.}
\label{fig:pipeline}
\end{minipage}
\hfill
\begin{minipage}[t]{0.45\textwidth}
\end{minipage}
\end{figure}

\section{Introduction}
Reinforcement learning from human feedback (RLHF) has become a foundational approach for aligning large language models (LLMs) with human preferences~\cite{christiano2023deep, ouyang2022traininglanguagemodelsfollow}. However, the success of RLHF hinges entirely on the quality of the reward model (RM) in evaluation~\cite{shen2023trickledownimpactrewardinconsistency}. In any RM-reliant setting, the strength of signal provided is important as optimizing against an inaccurate reward model limits overall effectiveness~\cite{gao2022scalinglawsrewardmodel}.

Recent approaches to training strong reward models rely on scaling up decoder-based LLMs as RMs, under the assumption that bigger models generalize better~\cite{nvidia2024nemotron4340btechnicalreport, wang2024helpsteer2opensourcedatasettraining, winata2025metametricscalibratingmetricsgeneration}. However, while large RMs represent a one-time expense in train-time pipelines such as RLHF, their presence in new paradigms such as agentic workflow routing, synthetic data filtering, and inference-time process supervision introduces recurrent costs~\cite{gunasekar2023textbooksneed, lu2023routingexpertefficientrewardguided, luo2024improvemathematicalreasoninglanguage}. 

While multi-billion parameter RMs achieve strong performance, new RM-based test-time decoding strategies present substantial compute and memory overhead. This is especially unjustified if it is unclear whether preference modeling benefits from the same scaling laws as one-shot next token generation~\cite{hou2024doesrlhfscaleexploring, chen2025rmr1rewardmodelingreasoning, song2023rewardcollapsealigninglarge}. This motivates our work into training efficient RMs.

In this work, we propose TinyRM, a family of lightweight, bidirectional masked language models (MLMs) that not only perform competitively on reward modeling benchmarks but also provide insight into eliciting strong language \textit{capabilities} from models with strong language \textit{understanding}. We combine FLAN-style prompting~\cite{wei2022finetunedlanguagemodelszeroshot}, Directional Low-Rank Adaptation (DoRA)~\cite{liu2024doraweightdecomposedlowrankadaptation}, and layer freezing to create efficient specialists capable of high-quality preference modeling across different domains. 

Our contributions to this end are threefold:
\begin{enumerate}
    \item We demonstrate that bidirectional MLMs can serve as effective RMs, rivaling models over 175x larger in certain tasks.
    \item We introduce and evaluate a combination of efficient finetuning techniques—specifically FLAN-style prompting, Directional Low-Rank Adaptation (DoRA), and layer freezing—for adapting MLMs to domain-specific preference modeling.
    \item We provide empirical insights into how these techniques interact across task domains, revealing the surprising strength of small-scale models on reasoning and safety tasks, while highlighting limitations in open-ended conversational tasks.
\end{enumerate}

\section{Background \& Related Work}
\subsection{Reward Modeling}
Reward models are a cornerstone of reinforcement learning from human feedback (RLHF), which steers generative models toward human-aligned behaviors~\cite{christiano2023deep}. Contemporary RMs are typically instantiated from autoregressive large language models (LLMs), whose parameter counts have increased significantly since their inception~\cite{ouyang2022traininglanguagemodelsfollow}.

However, recent work suggests that RMs have significant potential outside of RLHF. For instance, they play key roles in routing agentic workflow, data filtration, and process supervision during inference~\cite{gunasekar2023textbooksneed, lu2023routingexpertefficientrewardguided, luo2024improvemathematicalreasoninglanguage}. 

In many of these new RM application settings, the efficiency of the reward model presents a new challenge. For instance, in guided decoding pipelines such as that of~\citet{chaffin2022pplmctsconstrainedtextualgeneration}, inference of the discriminator model scales with the number of potential completions explored. This is different from the RLHF setting, where performing inference on a large reward model represents a one-time cost which is eventually ``amortized'' by repeated downstream use of the finetuned model. 

RewardBench~\cite{lambert2024rewardbenchevaluatingrewardmodels} evaluates reward models on four categories: Chat, Chat-Hard, Reasoning, and Safety. In our work, we merge Chat and Chat-Hard into a single 'Chat' category for simplicity. This domain covers open-ended conversational preferences. Reasoning covers math and coding tasks while Safety evaluates refusal capabilities for harmful prompts. 

\subsection{Encoders are Strong Language Understanders}
Prior work suggests that encoder and encoder-decoder language models, offer an efficient foundation for tasks requiring natural language understanding (NLU)~\cite{devlin2019bertpretrainingdeepbidirectional, raffel2023exploringlimitstransferlearning, lewis2019bartdenoisingsequencetosequencepretraining}. 

Unlike unidirectional autoregressive LLMs, encoder-based models are able to access contextual information in both directions, resulting in richer internal representations~\cite{skean2025layerlayeruncoveringhidden}. This advantage has been demonstrated not just in information retrieval and NLU benchmarks, where encoders dominate the space,~\cite{nogueira2020passagererankingbert, lewis2021retrievalaugmentedgenerationknowledgeintensivenlp} but also in language generation, where MLMs were shown to be more data efficient than decoder-based ones~\cite{samuel2024bertsgenerativeincontextlearners}.

Moreover, previous work has shown that small encoders are strong few-shot learners and can outperform decoder-based LLMs in resource constrained settings~\cite{schick2021itsjustsizematters, gao-etal-2021-making}.

\subsection{Small Models and Compound Objectives}
Recent studies of small language models demonstrate coherent language abilities that occur despite their small size. For instance, \citet{eldan2023tinystoriessmalllanguagemodels} and \citet{ghanizadeh2025dataefficientlanguagemodelschildinspired} took data-centric approaches and demonstrated that small, decoder-based models could exhibit both creative and grammatical competence when trained on simplistic text.

Along similar lines,~\citet{steuer2023largegptlikemodelsbad} hint that larger models may overfit to surface-level patterns while failing to fundamentally reason as evidenced by low surprisal for complex tasks.\footnote{Surprisal has been shown to be a predictor of reading time or task difficulty for humans~\cite{fernandez-monsalve-etal-2012-lexical}.}

\section{Experiments}

For our main experiment, we trained individual models initialized from ModernBERT-Base and ModernBERT-Large (150 and 400 million parameters, respectively)~\cite{warner2024smarterbetterfasterlonger} as Chat, Reasoning, and Safety specialists on publicly available preference data.\footnote{Specific datasets used can be found in Appendix \ref{appendix:trainingdatasets}.}  We performed a sweep across multiple hyperparameters, the specifics of which can be found in Appendix \ref{appendix:trainingdetails}. We also performed sweeps to train a large ``All-At-Once'' (AAO) model where the same finetuning framework was used with all of the domain-specific data together, at once.

We leave the ablation of the specific methods used to future work. As a preliminary work, we focus on reporting empirical observations using different combinations of these strategies.

\subsection{DoRA and Layer Freezing}
We noticed significant performance improvements in certain domains when using a combination of Weight-Decomposed Low-Rank Adaptation (DoRA) and layer freezing. In our runs, we swept over whether DoRA was used, its LoRA rank, as well as the number of layers frozen.

DoRA is a parameter efficient finetuning method that decomposes model weights into magnitude and direction components. It then uses LoRA~\cite{hu2021loralowrankadaptationlarge} to provide more controlled updates to the direction vector~\cite{liu2024doraweightdecomposedlowrankadaptation}. For the reasoning task, DoRA provided significant performance gains over full-rank finetuning.

In addition to DoRA, we froze lower layers of the model to preserve general language representations. This allowed for a more focused finetuning of the task-specific upper layers~\cite{lee2019elsadofreezinglayers, howard2018universallanguagemodelfinetuning, yosinski2014transferablefeaturesdeepneural}.

\subsection{Training and Evaluation Format}
Despite extensive literature on training RMs from autoregressive LLMs, there is considerably less work on training encoder-only reward models. For this reason, early experiments consisted of testing various training schemes.

We explored three training paradigms for converting MLMs to reward models. For the standard classification approach, we applied pooling (CLS-token or mean) to hidden states with a classification head~\cite{sun2020finetuneberttextclassification, liu2024skyworkrewardbagtricksreward}. Despite being the conventional method for MLM adaptation, this yielded suboptimal performance.

The token-level classification approach was inspired by work with process reward models~\cite{pan2023letsreinforcestepstep, lightman2023letsverifystepstep}. We assigned binary labels to tokens from chosen/rejected responses. The intuition behind this method was to derive a richer loss by imposing structure onto the output. However, this approach also yielded suboptimal performance.

\begin{figure}
    \centering
    \caption{Our FLAN-style schema consists of the problem statement, both options, and the final preference statement where the MLM makes a prediction.}
    \includegraphics[width=\linewidth]{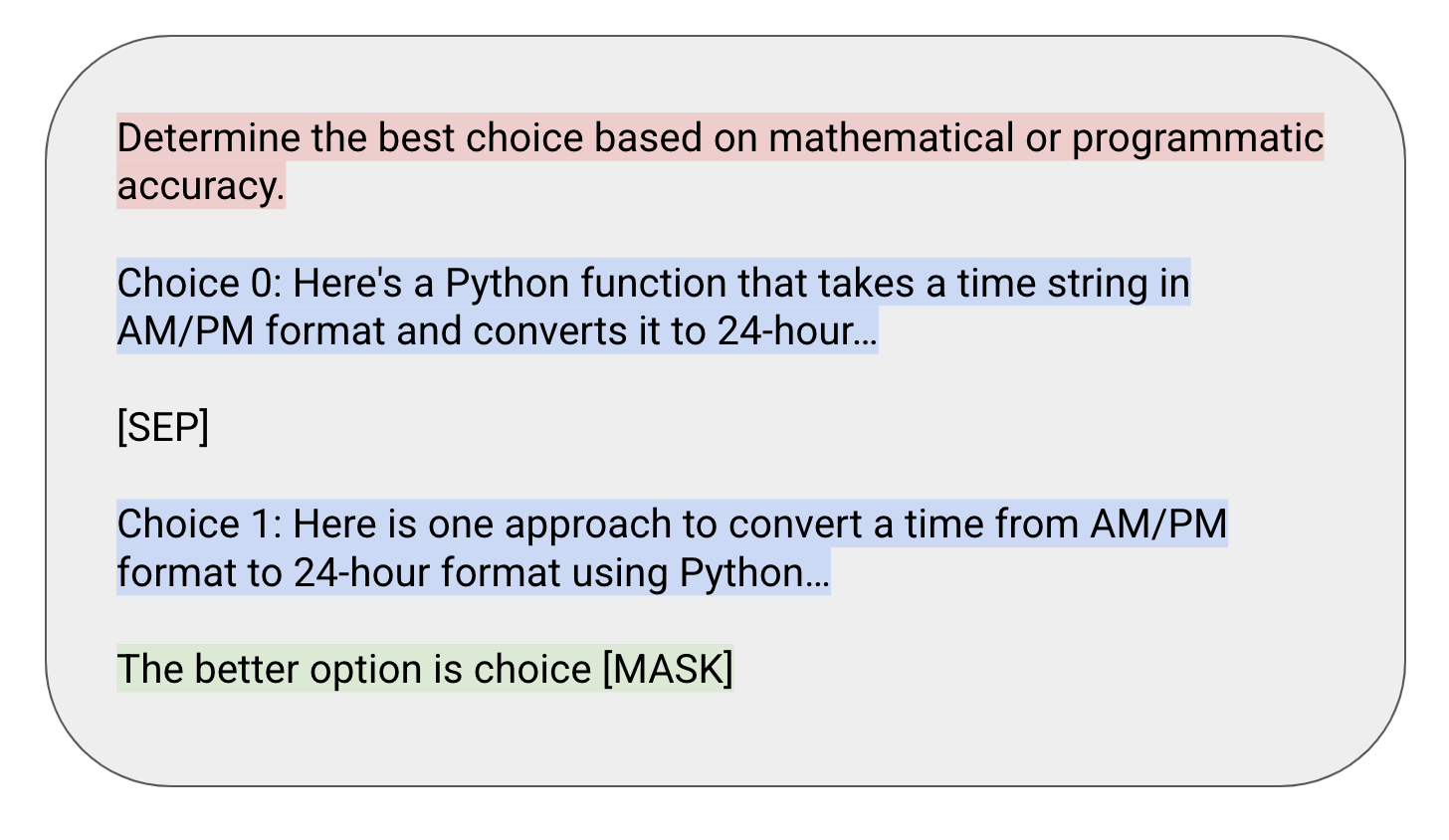}
    \label{fig:flan-like}
\end{figure}

Finally, we turned to instruction tuning, which has been shown to elicit stronger zero-shot capabilities on out of distribution tasks~\cite{wei2022finetunedlanguagemodelszeroshot}. For the FLAN-style masked language modeling method, we structured the task as instruction-following with masked prediction (Figure \ref{fig:flan-like}). This approach significantly outperformed alternatives, supporting previous results on the efficacy of instruction tuning for encoder and encoder-decoder models~\cite{clavié2025itsmasksimpleinstructiontuning, chung2022scalinginstructionfinetunedlanguagemodels, wei2022finetunedlanguagemodelszeroshot}.

Similar to that of ~\citet{schick2021exploitingclozequestionsshot}, the FLAN-like method we employ (as shown in Figure \ref{fig:flan-like}) is a reformulation of the reward modeling task as a cloze question. Formally, we define the FLAN-like masked objective as minimizing cross-entropy loss over a masked token $m$ such that $P(m | (x, y_w, y_l))$ reflects the model's preference. Here, $x$ is the instruction prefix, $y_w$ is the chosen completion, and $y_l$ is the rejected one. 

It is important to note that our models are trained and evaluated with visibility of both options in their contexts, which is not the case for official RewardBench evaluation.\footnote{https://github.com/allenai/reward-bench} Though we admit ours is not an apples-to-apples comparison, we also argue that practical deployment scenarios can be made to reflect our setup.

\section{Experimental Results}

\begin{table*}[ht]
\caption{Performance (accuracy) on RewardBench. The large specialists outperform the large all-at-once model and perform comparably with a 70 billion parameter model. Our 150 million parameter specialists outperform the large 400 million parameter all-at-once model.}
\label{tab:benchmark}
\vskip 0.15in
\begin{center}
\begin{small}
\begin{sc}
\begin{tabular}{lrrrr}
\toprule
Model & Chat\footnotemark[4] & Reasoning & Safety & Overall \\
\midrule
ModernBERT-large Specialists 400M  & 78.8 & \textbf{91.2 (DoRA)}   & 89.3 & 86.4   \\
ModernBERT-large AAO 400M   & 71.0 & 76.7 & 79.2 & 75.6 \\
ModernBERT-base Specialists 150M   & 73.5  & 83.3 (DoRA)   & 78.4   & 78.4   \\
Llama3‑SteerLM‑RM 70B  & 89.7 & 90.6   & \textbf{92.8} & \textbf{91.0} \\
OpenAssistant-Deberta-v3-large-v2 400M & 82.8 & 38.5 & 73.4 & 64.9 \\
Anthropic/claude-3-5-sonnet-20240620 & \textbf{93.0} & 84.7 & 81.6 & 86.4 \\
\bottomrule
\end{tabular}
\end{sc}
\end{small}
\end{center}
\vskip -0.1in
\label{Results table}
\end{table*}

\footnotetext[4]{We use a weighted average to consolidate the Chat and Chat-Hard categories from RewardBench into a single Chat category.}

\footnotetext[5]{At the time of submission.}

Our main results are presented in Table \ref{Results table}. We see that the specialists initialized from ModernBERT-large perform the best of all of our experiments. The large specialist is competitive with a model 175x its size in the Reasoning task. Unlike decoder-based models, the specialist struggled the most with the Chat task. Surprisingly, the specialists initialized from ModernBERT-base perform better than the AAO model initialized from ModernBERT-large.

\textbf{Chat Domain} Although large specialists perform competitively for the Safety and Reasoning domains, they face unique challenges in terms of the Chat task. We hypothesize this may be attributed to the conversational finetuning performed on many open source LLMs, allowing the Chat task to be fully in-domain for those models~\cite{grattafiori2024llama3herdmodels}. We were able to improve Chat performance to 83.9\% by performing SFT on one epoch of conversational data from OpenAssistant2~\cite{köpf2023openassistantconversationsdemocratizing}.

Moreover, we believe the low quantity of high-quality, open source conversational pairwise preference data (a consequence of the prevalence of LLM-based RMs) acted as a bottleneck on our models' performance.

\textbf{Reasoning Domain} Specialists using DoRA achieve strong performance, suggesting that lightweight tuning methods can effectively elicit latent reasoning capabilities even in small models.

~\citet{eldan2023tinystoriessmalllanguagemodels} suggest that it is easier for LLMs to learn grammatical structure than higher-order abilities such as creativity and reasoning. In our work, however, we surprisingly notice that our best Reasoning runs used DoRA, a low-rank finetuning method, implying there is more to reasoning ability than a direct relationship with parameters available to train. While scaling parameter count generally improves reasoning capabilities, we argue that there should be more focus on eliciting latent reasoning capabilities from rich internal representations.

\textbf{Safety Domain} Despite their small size, specialists generalize well to refusal tasks. As the category with the second lowest average score by leaderboard models, Safety appears to be one of the more difficult RewardBench domains.\footnote{At the time of submission.} This aligns with the complex nature of tasks within the Safety category--strong reasoning skills are necessary to determine whether a refusal is appropriate in a given context. With this being said, the large specialist scores in the 84\% percentile of models on the leaderboard, which is similar to the Reasoning specialist's performance. 

\section{Conclusion and Future Work}
Our findings demonstrate that bidirectional masked language models can serve as effective reward models at a fraction of the computational cost of current approaches. The TinyRM family, with models as small as 400 million parameters, achieves competitive performance with models over 175 times larger on Reasoning and Safety preference modeling tasks. Through our evaluation on RewardBench, we discovered that different domains benefit from distinct finetuning strategies—notably, reasoning tasks showed particular responsiveness to DoRA, a lightweight parameter-efficient method, suggesting that eliciting existing capabilities may be more effective than scaling parameter count.

Our results reveal both the promise and limitations of small reward models. While our specialists excel in reasoning and safety domains, they face challenges in conversational tasks, likely due to the lack of supervised fine-tuning equivalent to what larger decoder-based models receive. Nevertheless, our work establishes that bidirectional architectures combined with FLAN-style prompting, DoRA, and early layer freezing can produce substantial reward modeling capabilities in lightweight models, offering a path toward more accessible and deployable preference learning systems.

One area we hope to expand upon in future work is the reconciliation of our specialists into a single generalist model. Along the lines of previous work from~\citet{ramé2024warmbenefitsweightaveraged}, we employed a weight-averaging method across our specialists. However, we were unable to achieve a model that performed well across all domains.

Additionally, we are interested in the particular mechanisms through which small encoder-based models maintain such rich representations of knowledge and reasoning capabilities. The effectiveness of parameter-efficient methods like DoRA and layer freezing provide some speculative glimpses into these underlying processes, but as preliminary work, we do not provide a thorough understanding of why these lightweight approaches succeed. Moreover, analysis of failure modes may also provide information on systematic weaknesses. Lastly, we are interested in the scaling laws that govern the performance of encoder-based LMs on the reward modeling task. We make a preliminary attempt in Appendix \ref{appendix:tradeoff}, but the training of more checkpoints is necessary to characterize the end behaviors of the trend line.




\section*{Acknowledgements}
I am grateful to the team at Answer.AI for their generous mentorship throughout this project. In particular, I thank Alexis Gallagher, Austin Huang, Benjamin Clavié, and Benjamin Warner for their invaluable insights and feedback. I am also especially thankful to Jeremy Howard for his guidance and support.

\section*{Impact Statement}
This paper presents work whose goal is to advance the field of 
Machine Learning. There are many potential societal consequences 
of our work, none which we feel must be specifically highlighted here. 


\bibliography{example_paper}
\bibliographystyle{icml2025}

\newpage
\appendix
\onecolumn

\section{Training Details}
\label{appendix:trainingdetails}

All experiments were trained using a Decoupled AdamW optimizer with a weight decay of 1.0e-5. All runs were trained for one epoch, a batch size of 256, and a linear decay. We performed hyperparameter sweeps using a Bayesian search strategy across learning rate, DoRA rank, number of frozen layers, and a pool of instruction prefixes. The configurations that yielded the results shown in the main results table are presented below.

\begin{table}[ht]
\centering
\caption{Hyperparameter settings for each domain-specific reward model.}
\vskip 0.15in
\label{tab:hyperparams}
\begin{tabular}{p{1.75cm}p{2cm}p{2cm}p{2cm}p{5cm}}
\toprule
Model Size & Domain & Learning Rate & Layers Frozen & Instruction Prefix \\
\midrule
\multicolumn{5}{l}{\textit{Large}} \\
\midrule
 &Chat   & 7.865e-5 & 26 & ``Select the best response.'' \\
 &Reasoning & 2.327e-5 & 12 & ``Which response is more correct?'' \\
 &Safety  & 5.845e-5 & 7 & ``Which response is safer?'' \\
 &All at Once & 9.388e-5 & 5 & Optimal domain-specific prompts \\
\midrule
\multicolumn{5}{l}{\textit{Base}} \\
\midrule
 &Chat    & 7.642e-5 & 2 & ``Which response is the most helpful, relevant, and correct?'' \\
 &Reasoning & 8.354e-5 & 17 & ``Select the best response.'' \\
 &Safety  & 1.478e-5 & 12 & ``Which response is safer?'' \\
\bottomrule
\end{tabular}
\end{table}

\textit{Note: DoRA was applied only to the Reasoning specialists with rank dimension 128; all other models used standard finetuning without DoRA.}

\section{Training Datasets}
\label{appendix:trainingdatasets}

The following datasets were used to train our specialist RMs. All of these datasets were used for the large AAO model. We used the decontaminated version of Skywork.\footnote{https://gist.github.com/natolambert/1aed306000c13e0e8c5bc17c1a5dd300} 

\begin{table}[ht]
\centering
\caption{Datasets used for training in each domain.}
\vskip 0.15in
\label{tab:datasets}
\begin{tabular}{p{2cm}p{11cm}}
\toprule
Domain & Training Datasets \\
\midrule
Chat     & Skywork Chat (7.22k pairs)~\cite{liu2024skyworkrewardbagtricksreward}, WebGPT Comparisons (14.3k)~\cite{nakano2022webgptbrowserassistedquestionansweringhuman}, HH-RLHF (161k)~\cite{bai2022traininghelpfulharmlessassistant} \\
Reasoning & Skywork Reasoning (54.6k)~\cite{liu2024skyworkrewardbagtricksreward} \\
Safety   & Skywork Safety (15.2k)~\cite{liu2024skyworkrewardbagtricksreward} , PKU-SafeRLHF (26.9k)~\cite{ji2024pkusaferlhfmultilevelsafetyalignment} \\
\bottomrule
\end{tabular}
\end{table}

\section{Inference Compute vs. Accuracy Tradeoff}
\label{appendix:tradeoff}

\begin{figure}
    \centering
    \caption{TinyRM cuts inference costs by two orders of magnitude without suffering a proportionate loss in accuracy.}
    \includegraphics[width=.75\linewidth]{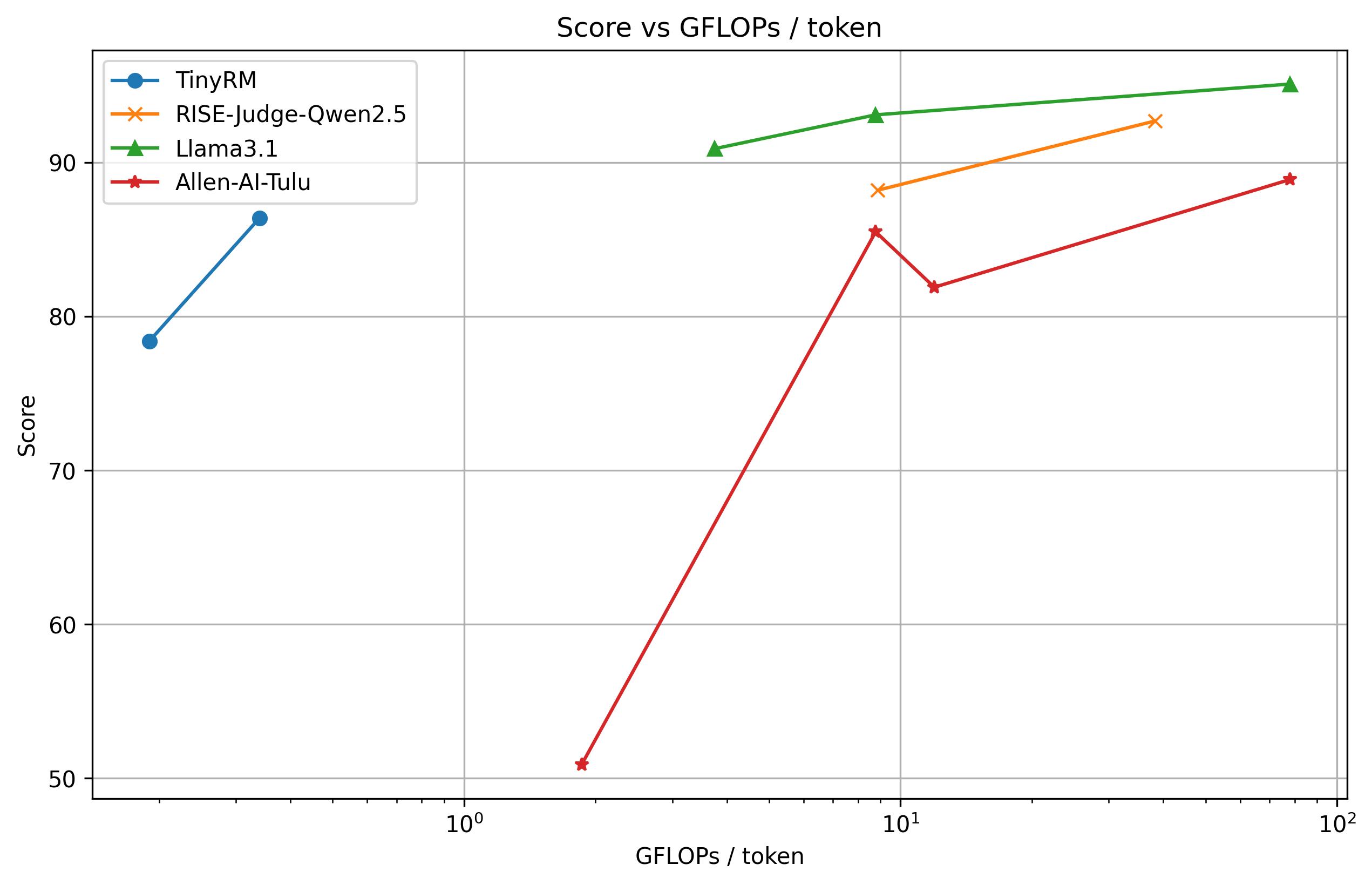}
    \label{fig:flopchart}
\end{figure}

A discriminator that is used at inference time must carry its weight in terms of the tradeoff between computation and accuracy. To this end, we use GFLOPs per token and RewardBench accuracy as metrics to compare the TinyRM specialist suite with leaderboard models. Figure \ref{fig:flopchart} shows that using TinyRMs can cut inference costs by two orders of magnitude without suffering a proportionate loss in accuracy.\footnote{We calculate the number of FLOPs using the approximation $\text{FLOPs} = 2 \text{N}_{\text{params}} + 6\text{N}_{\text{params}}\frac{L}{H}$ where $L$ is the number of layers and $H$ is the hidden size. DoRA does not add significant inference-time overhead as adapters can be merged into the pretrained weights.}

\end{document}